\pdfoutput=1
\documentclass[11pt]{article}

\usepackage{ACL2023}

\usepackage{times}
\usepackage{latexsym}

\usepackage[T1]{fontenc}

\usepackage[utf8]{inputenc}

\usepackage{microtype}

\usepackage{inconsolata}

\usepackage{amsfonts}
\usepackage[fleqn]{amsmath}
\usepackage{hyperref}
\usepackage{graphicx}
\usepackage{placeins}
\usepackage{subcaption}
\usepackage{tikz}
\usetikzlibrary{decorations.pathreplacing}
\DeclareMathOperator*{\argmin}{argmin}
\DeclareMathOperator{\halfnorm}{\text{\TH}}
\newcommand\given[1][]{\:#1\vert\:}
\newcommand\nexta{\frac{a_{j} + a_{j + 1}}{2}}
\newcommand\preva{\frac{a_{j - 1} + a_{j}}{2}}

\title{NF4 Isn't Information Theoretically Optimal (and that's Good)}

\author{Davis Yoshida\\Toyota Technological Institute at Chicago, IL, USA, 60637\\\texttt{dyoshida@ttic.edu}}

\begin{document}
\maketitle
\begin{abstract}
This note shares some simple calculations and experiments related to absmax-based blockwise quantization, as used in \citet{qlora}.
Their proposed NF4 data type is said to be information theoretically optimal for representing normally distributed weights.
I show that this can't quite be the case, as the distribution of the values to be quantized depends on the block-size.
I attempt to apply these insights to derive an improved code based on minimizing the expected L1 reconstruction error, rather than the quantile based method.
This leads to improved performance for larger quantization block sizes, while both codes perform similarly at smaller block sizes.
\end{abstract}

\section{Introduction}

\begin{figure}[h!]
\includegraphics[width=0.9\columnwidth]{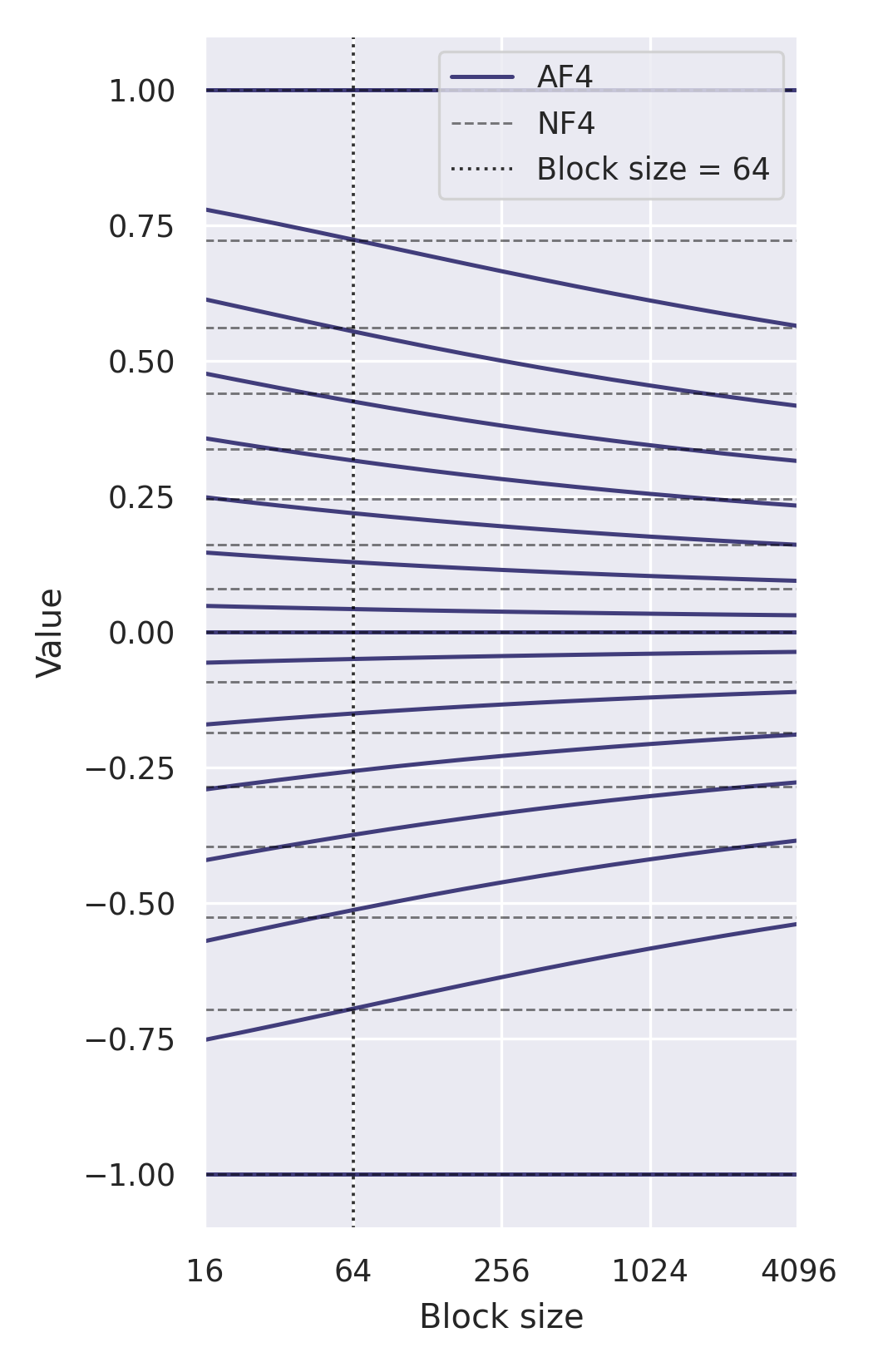}
\caption{For varying $B$, the 16 values of AF4-$B$, a code based on minimizing expected L1 quantization error. At higher block sizes the values to be quantized will cluster more tightly around the origin, so the code values should as well. Horizontal lines are the NF4~\citep{qlora} code values, the vertical line shows the default NF4 block size of 64.}
\end{figure}\label{fig:codes}

This note will go over some theoretical calculations and empirical findings regarding quantization using the NF4 data type of \citet{qlora}. I want to stress that while I am pointing out some theoretical issues, they don't have any bearing on the \emph{empirical} results of ~\citet{qlora}, since all that matters at the end of day is how well a model performs post-quantization.

A brief summary of my findings is: (1) The distribution of values to be quantized depends on the quantization block size, so an optimal code should vary with block size (2) NF4 does not assign an equal proportion of inputs to each code value (3) Codes which \emph{do} have that property are not as good as NF4 for quantizing language models.

In Section~\ref{af4}, I propose a code (shown in Figure~\ref{fig:codes}), which takes these facts into account, but it only consistently outperforms NF4 for larger block sizes in preliminary experiments.
\section{Background}\label{background}
The NF4 datatype is used to quantize values in the interval $[-1, 1]$.
Since NF4 is a 4-bit code, it consists of 16 values, $q_1, \dots, q_{16} \in [-1, 1]$.
Each block of $B$ values of a matrix $W$ is quantized into NF4 values:
\begin{enumerate}
    \item Given $w_1, \dots, w_B$, calculate the \emph{absmax}: $M = \max\limits_{i} |w_i|$
    \item Calculate a code index, $c_i$ for each parameter by mapping each $w_i$ to the nearest $q_j$ after downscaling by $M$:
    \begin{equation*}
        c_{i} = \argmin\limits_{j} |q_j - w_i/M|
    \end{equation*}
\end{enumerate}
In addition to the 4-bit NF4 values, the absmax $M$ is stored for each block.
When $W$ is used in a computation, blocks are dequantized by computing $w_i \approx q_{c_i} M$.

\citet{qlora} show that neural network parameters are approximately normally distributed.
Motivated by this, they calculate the values of $q_j$ based on the quantiles of the normal distribution.
Their construction is:\footnote{The implementation can been found \href{https://github.com/TimDettmers/bitsandbytes/blob/0f40fa3f0a198802056e29ba183eaabc6751d565/bitsandbytes/functional.py\#LL236C5-L236C22}{here} in the \texttt{create\_normal\_map} function.}
\begin{enumerate}
    \item Set $\delta = \frac{1}{2}\left(\frac{1}{32} + \frac{1}{30}\right)$.
    \item Compute 8 evenly spaced probability values $p_1, \dots, p_8$ such that $p_1 = \delta$ and $p_8 = 1/2$.
    \item Find their pre-images under the Gaussian CDF, $\Phi$: $\tilde{q}_i = \Phi^{-1}(p_i)$ for $i = 0, \dots, 8$. 
    \item Compute 9 even spaced probability values $r_8,\dots,r_{16}$ such that $r_8 = 1/2$ and $r_{16} = 1 - \delta$.
    \item Set $\tilde{q}_i = \Phi^{-1}(r_i)$ for $i = 9, \dots, 16$ (note that $r_8$ is unused since $\tilde{q}_8$ was already set to 0)
    \item Normalize the $\tilde{q}$s to the range $[-1, 1]$ to get the final code: $q_i = \frac{\tilde{q}_i}{\max\limits_{i} |\tilde{q}_i|}$.
\end{enumerate}
This results the NF4 code: a set of values in $[-1, 1]$, with $q_1 = -1$, $q_8 = 0$, $q_{16} = 1$.\footnote{The reason the the values are computed in two asymmetric groups is that 0 would not be included in a symmetric code.}
The largest $\tilde{q}$ value is $\Phi^{-1}(1 - \delta) \approx 1.848$, so the final $q$ values are quantiles of an $\mathcal{N}\left(0, \frac{1}{\Phi^{-1}(1 - \delta)}\right)$ distribution.

\section{Increased block size concentrates the inputs}
\begin{figure*}[t]
    \centering
    \includegraphics[width=\textwidth]{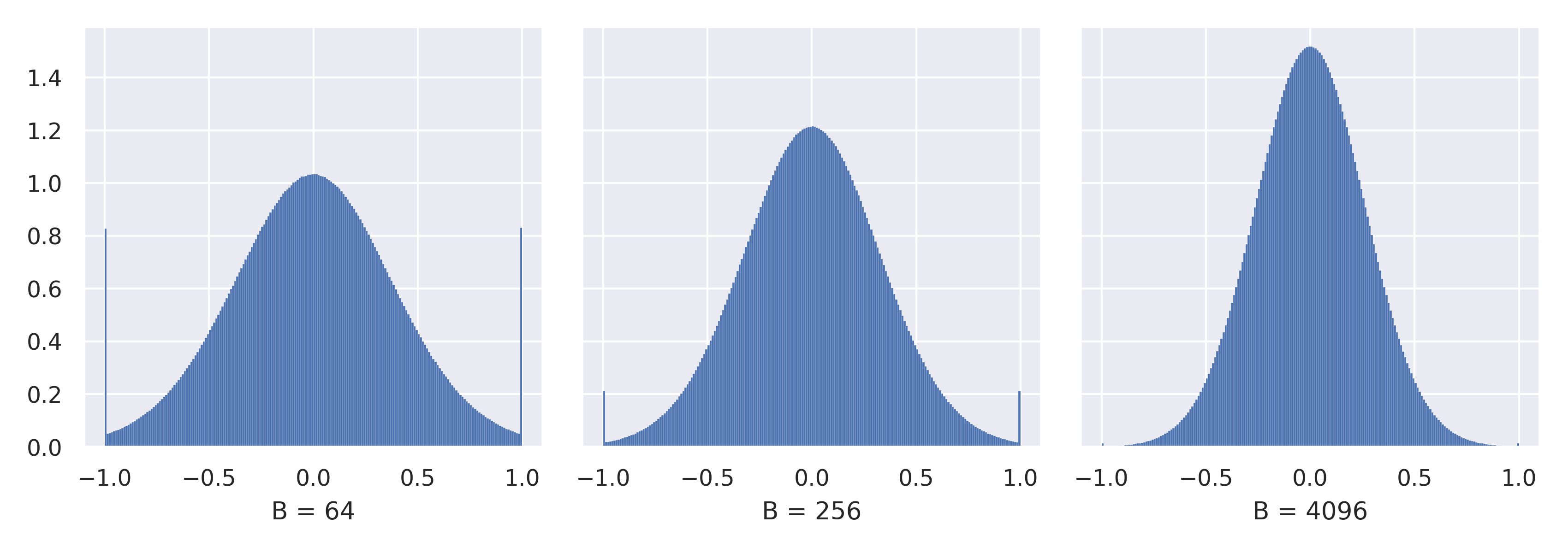}
    \caption{Estimates of the density of $X_i$ for varying values of $B$. Each plot uses $2^{20}$ draws of $B$ values.}\label{fig:dists}
\end{figure*}
To allow theoretical analysis of this quantization method, consider the following generative story for $X_1, \dots, X_B$:
\begin{align}
Z_1,\dots,Z_B &\sim\mathcal{N}(0,1)\nonumber\\
M &= \max\limits_{i} |Z_i|\nonumber\\
X_i &= \frac{Z_i}{M}\label{eq:genproc}
\end{align}
This models quantizing samples from a normal distribution in groups of $B$.
The $X_i$ are identically distributed, but there is dependence introduced by the division by $M$.\footnote{E.g. $P(X_1{=}1) = 1/2B$, but $P\left[X_1 = 1 \given[\middle] X_2 = 1\right] = 0$}

\subsection{One code cannot be optimal for all block sizes}
The 16 NF4 codepoints do not depend on $B$, but the distribution of the $X_i$s above does. Figure~\ref{fig:dists} shows estimates of the density of this distribution for various values of $B$.
The distribution is a mixed distribution, since with probability 1 exactly one $X_i$ obtains either the value -1 or 1.
So, each $X_i$ (non-independently) has a probability of $1/2B$ of falling on one of those two values.
More importantly, the distribution concentrates around the origin as $B$ is increased.

The relationship of this to the selection of code values can be seen through a brief example.
Consider using a block size of 4096, which corresponds to one row of an attention projection matrix in the LLaMA-7B model~\citep{llama}.
We can show that the larger code values such as $q_{15}$ will barely be used at all.

Let $\halfnorm(\cdot)$ be the CDF of the half-normal distribution\footnote{``\TH'' is the chracter ``thorn''}.
$M$ has CDF $\halfnorm^{4096}(\cdot)$, which may be used to compute its median, $m_B$:
\begin{align*}
   m_B &= \halfnorm^{-1}\left(\left(\frac{1}{2}\right)^{1/4096}\right)\\
   &\approx 3.76
\end{align*}

The largest value in $[-1, 1]$ which is assigned to $q_{14}$ in the NF4 code is slightly smaller than 0.65. When $M$ attains its median value, the fraction of samples assigned to $q_{15}$ and $q_{16}$ will be:
\begin{align*}
    P&\left[X_i > 0.65 \given[\middle] M = 3.76, X_i \neq M\right]\\
    &=P\left[Z_i > \left(0.65\right)\left(3.76\right) \given[\middle] -3.76 < Z_i < 3.76\right]\\
    &= \frac{P\left[\left(0.65\right)\left(3.76\right) < Z_i < 3.76\right]}{P(-3.76 < Z_i < 3.76)}\\
    &= \frac{\Phi(3.76) - \Phi\left((0.65)(3.76)\right)}{\Phi(3.76) - \Phi(-3.76)}\\
    &\approx 0.007\\
\end{align*}
So when using a block size of 4096, the expected number of samples assigned to code values $q_{15}$ and $q_{16}$ will be less than 1\%, whenever $M$ obtains its median value or higher.

One can generalize the above calculation to show that for any fixed code, a large enough block size will lead all but the smallest code values being used extremely infrequently.

\subsection{The exact distribution of $X_i$}
As mentioned earlier, $X_i$ follows a mixture distribution, with mass $1/B$ being assigned uniformly to $\{-1, 1\}$, and mass $(B-1)/B$ spread across $(0, 1)$.
I'm keeping the index on $X_i$ to remind the reader that it comes from a sample which contains other values of which it is not independent.
To avoid needing to carry all these $B$'s around, I'll condition on $|X_i| \neq 1$, i.e. the corresponding $Z_i$ is not the maximal sample in the block.

For a fixed value of $M$, the conditional CDF is:
\begin{align}
    &P\left[X_i \le x \given[\middle] |X_i| < 1, M=m\right]\nonumber\\
    &= P\left[Z_i < mx \given[\middle] |Z_i| < m\right]\nonumber\\
    &=\frac{\Phi(mx) - \Phi(-m)}{\Phi(m) - \Phi(-m)}\nonumber\\
    &= \Psi(mx;m,1)\label{eq:cond_cdf}
\end{align}
where $\Psi(\cdot;m,\sigma)$ is the CDF of a truncated normal distribution with limits $[-M, M]$, mean $0$, and standard deviation $\sigma$.

To compute integrate out the conditioning on $M$, we need to calculate its PDF, $p_m$. Differentiating its CDF, $F_M$:
\begin{align*}
F_M(m) &= \halfnorm^B \left(m\right)\\
p_m(m) &= 2B \halfnorm^{(B-1)} \left(m\right) \varphi(m)
\end{align*}
where $\varphi(\cdot)$ is the Gaussian PDF.

Combining this expression with Equation~\ref{eq:cond_cdf}, define $G_B$ to be the CDF of the continuous portion of $X_i$'s distribution:
\begin{align*}
    &P\left[X_i < x \given[\middle] |X_i| < 1\right]\\
    &= G_B(x)\\
    &= 2B \int_0^\infty \halfnorm^{(B-1)} \left(m\right) \varphi(m) \Psi\left(x;m,1\right) dm\\
\end{align*}
There's a good approximation of $G_B$ (See Appendix\ref{approximation}), but to avoid worrying about the quality of approximation, I'll just use numerical integration to calculate it directly.

Combining this with the discrete distribution for the endpoints, the CDF of $X_i$ with block size $B$, $F_X(x;B)$, is:
\begin{align}
    F_X(x&;B) = P\left[X_i \le x\right]\nonumber\\
    &= \begin{cases}
        0 & x < -1\\
        \frac{1}{2B} & x = -1\\
        \frac{1}{2B} + G_B(x) & -1 < x < 1\\
        1 & x \ge 1
    \end{cases}\label{eq:cdf}
\end{align}

\section{Quibbles about quantiles}
\citet{qlora} say that the defining property of NF4 is that it ``has equal expected number of values in each quantization bin.''.
I'll show in this section that this is not the case, and how to construct a code with that property.
However, it turns out that such codes generally perform worse than $NF4$, implying that one should not use this as a criterion for constructing codes for quantization (See Appendix~\ref{uniform}).

There is one minor ambiguity which needs to be cleared up before moving forward. In \citet{qlora}, the method given for constructing NF4 is to compute the average of quantile values, e.g.:
\begin{equation*}
    \tilde{q}_{1} = \frac{1}{2}\left(\Phi^{-1}(1/16) + \Phi^{-1}(2/16)\right)
\end{equation*}
On the other hand, the method used in the implementation is to compute the quantile function applied to averaged inputs:
\begin{equation*}
    \tilde{q}_{1} = \Phi^{-1}\left(\frac{1}{2}\left(1/16 + 2/16\right)\right)
\end{equation*}
These values are not equal because $\Phi^{-1}$ is nonlinear.
However, the final outcome differs by less than 0.001, so it is unlikely to have a noticeable effect.

\definecolor{color0}{rgb}{0.00392156862745098, 0.45098039215686275, 0.6980392156862745}
\definecolor{color1}{rgb}{0.8705882352941177, 0.5607843137254902, 0.0196078431372549}
\definecolor{color2}{rgb}{0.00784313725490196, 0.6196078431372549, 0.45098039215686275}
\definecolor{color3}{rgb}{0.8352941176470589, 0.3686274509803922, 0.0}
\definecolor{color4}{rgb}{0.8, 0.47058823529411764, 0.7372549019607844}
\definecolor{color5}{rgb}{0.792156862745098, 0.5686274509803921, 0.3803921568627451}
\definecolor{color6}{rgb}{0.984313725490196, 0.6862745098039216, 0.8941176470588236}
\definecolor{color7}{rgb}{0.5803921568627451, 0.5803921568627451, 0.5803921568627451}
\definecolor{color8}{rgb}{0.9254901960784314, 0.8823529411764706, 0.2}
\definecolor{color9}{rgb}{0.33725490196078434, 0.7058823529411765, 0.9137254901960784}
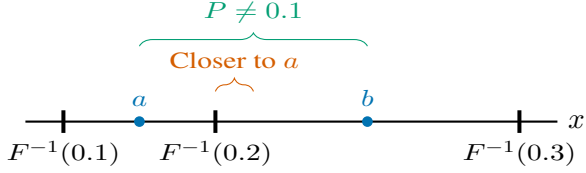
\begin{figure}
\begin{tikzpicture}

% Define colors
% x-axis
\draw[-, thick] (-0.5,0) -- (6.5,0) node[right] {$x$};

% Bold lines and probabilities at 0, 2, and 6
\foreach \x/\p in {0/0.1, 2/0.2, 6/0.3}
{
    \draw[line width=1.5pt] (\x,-0.15) -- (\x,0.15);
    \node[below,yscale=0.8] at (\x, -0.15) {$F^{-1}(\p)$};
}

% Points at 1 and 4 in color 1
\foreach \x/\n in {1/a,4/b}
{
    \fill[color0] (\x,0) circle (2pt);
    \node[color0,above=0.1cm,yscale=0.8] at (\x, 0) {$\n$};
}

\draw[decorate,decoration={brace,amplitude=5pt},yscale=1.0,color=color3] (2,0.4) -- (2.5,0.4) node[midway,above=0.2cm,yscale=0.8] {Closer to $a$};

% Curly brace in color1 and label "P ≠ 0.1" (if required)
\draw[decorate,decoration={brace,amplitude=5pt},yscale=1.0,color=color2] (1,1) -- (4,1) node[midway,above=0.2cm,yscale=0.8] {$P \neq 0.1$};
\end{tikzpicture}
\caption{Two bins of probability mass $0.1$ for a distribution with CDF $F$. \textcolor{color3}{They have unequal widths, so using their centers, $a$ and $b$, as code values will lead to too much mass being assigned to $a$.} \textcolor{color2}{If $F^{-1}(0.2)$ itself is used as a code value, the mass it encodes will depend on the distribution of mass within the bins to either side.}}\label{fig:bins}
\end{figure}
\begin{figure}
    \centering
    \includegraphics[width=\columnwidth]{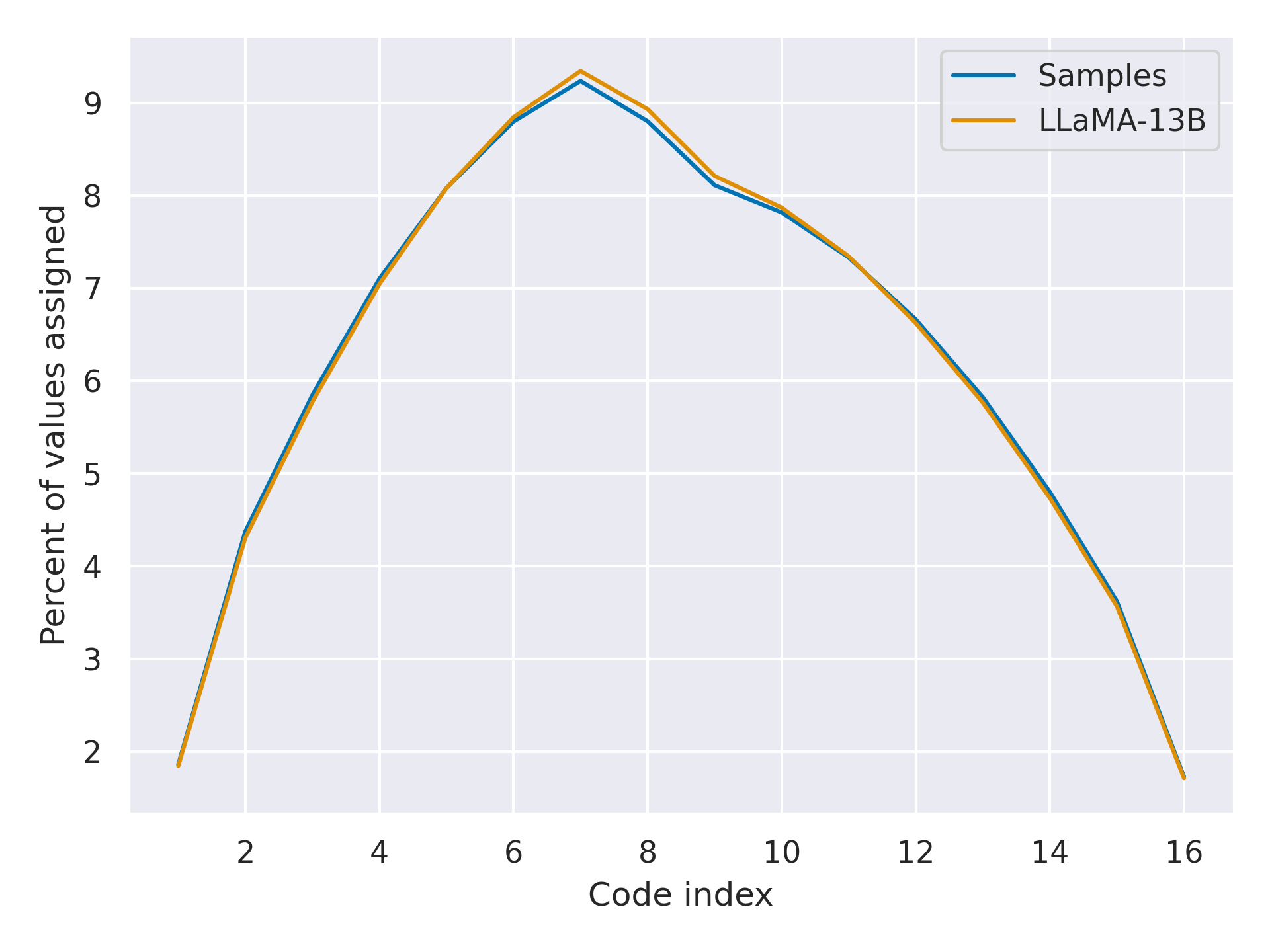}
    \caption{Usage of each NF4 code value for (a) Samples from the generative process in Equation~\ref{eq:genproc} (b) Weights from the LLaMA-13B model. (Quantization block size $B=64$)}
    \label{fig:usage}
\end{figure}
Regardless, neither of these strategies leads to a code which allocates equal probability mass to all quantization bins.
Figure~\ref{fig:bins} demonstrates the general issue, namely that non-linearity in the CDF will cause quantile-based methods to miss the mark.

Figure~\ref{fig:usage} shows the actual usage of NF4 code values for weights from LLaMA-13B, and samples from the process defined in Equation~\ref{eq:genproc}.
Rather than each value being used with probability 6.25\%, usages range between 2\% and 9\%.

\subsection{Constructing uniformly used codes}\label{construct_uniform}
Here's a way to construct a code which has bins $[b_1, b_2]$, $[b_2, b_3]$, etc. ($b_1$ may be $-\infty$):
\begin{enumerate}
    \item Choose an initial code value $q_1 \in [b_1, b_2]$.
    \item For $k=2, 3, \dots k_{\text{max}}$:
    \begin{equation*}
    q_k \gets 2 b_k - q_{k-1}
    \end{equation*}
\end{enumerate}
This will ensure that $(q_k + q_{k+1})/2 = b_{k+1}$, i.e. $b_{k+1}$ will be a quantization bin endpoint.
(Because $q_1$ was free to vary, this describes a set of codes rather than a single unique code).

In preliminary experiments, I found that these codes performed quite badly for quantizing actual neural networks (See Appendix~\ref{uniform}).
This indicates that uniform usage of code values is not the idea target to aim for.

\subsection{Minimizing reconstruction error}\label{l1}
A single quantization error in a matrix causes a change in the output of multiplications by that matrix proportional to the size of the error.
Because of this, directly minimizing the absolute value of the expected quantization error seems like a promising approach.\footnote{This is a bit of a post hoc justification. I initially tried minimizing the expected squared error, but that led to worse LM performance.}

Given a random variable $Y$ which is to be quantized, this we are searching for the values $a_1 < a_2 < \dots < a_{16}$ which solve the following problem:
\begin{equation*}
    \min\limits_{a_1, \dots, a_{16}} E\left[\min_j\left\vert Y - a_j\right\vert\right]
\end{equation*}
This is a 1-dimensional k-medians problem, over a mixed probability distribution rather than a finite sample.

If $Y$ is continuous, it's easy to calculate the following stationarity condition for the optimal $j$-th code point for $1 < j < 16$:
\begin{align}
P\biggl[\preva &< Y < a_j \biggl]\nonumber\\
&=\label{eq:starionarity}\\
P\biggl[a_j < Y <& \nexta\biggl]\nonumber
\end{align}
This condition simply says that each code point should be the median of the values mapped to it.
It is a generalization of the well-known fact that the median of a univariate distribution minimizes the expected distance from a sample.

This constraint allows one to derive the value of $a_{j+1}$ from $a_j$ and $a_{j-1}$ as follows:
\begin{align*}
    F_Y&(a_j) - F_Y\left(\preva\right) =\\
    &F_Y\left(\nexta\right) - F_Y(a_j)
\end{align*}
So,
\begin{equation}\label{eq:step}
    a_{j+1} = 2F_Y^{-1} \left(\rho_j\right) - a_j 
\end{equation}
where
\begin{equation*}
    \rho_j = 2F_Y(a_j) - F_Y\left(\preva\right)
\end{equation*}
Given any two consecutive code values, the remainder of the values may be determined using this rule.
If we instead of two non-consecutive code values such as $a_1 = -1$ and $a_8 = 0$ (as in NF4), we can search for a value of $a_2$ such that $a_7$ is stationary:
\begin{align}
a_8 = 2F^{-1}_X (\rho_7) - a_7
\end{align}
$a_{9:15} \in (0, 1)$ may be found analogously if $a_{16} = 1$ is fixed as well.

\section{AF4: 4-bit AbnormalFloat}\label{af4}
In order to see whether the above calculations are of more than theoretical interest, I apply the method from the previous section to the $F_X(x;B)$ as defined in Equation~\ref{eq:cdf}.
The result is a code which I will call the \textbf{4-bit AbnormalFloat datatype with block size $B$} (AF4-$B$).
The three fixed values used for the construction are: $a_1 = -1$, $a_8 = 0$, and $a_{16} = 1$, which are also included in NF4.
A script for producing these codes is available at \url{https://github.com/davisyoshida/abnormal-floats}.

Figure~\ref{fig:codes} shows the result of using this process to produce AF4-$B$ codes for a variety of block sizes.
As expected, higher values of $B$ lead to codes more tightly clustered around 0 as the distribution of $X_i$ becomes more peaked.

One thing which I should emphasize is that AF4 is \emph{not} the code which globally minimizes the expected reconstruction error.
Requiring the code to include -1, 0, and 1 makes the average reconstruction error worse, but including these values in the code is essential for quantization performance.

An interesting point is that the outermost NF4 values happen to nearly coincide with AF4-64, and 64 is the default block size for using NF4.
The combination of this, the importance of including -1/0/+1, and the fact that uniform binning codes perform poorly, suggests that representing these larger values is important even if they occur less frequently.

\section{Experiments}
In order to check whether the reasoning above might be of more than theoretical interest, I have run some experiments on language modeling and zero-shot classification.
The high-level summary is that the AF4 code construction method does not consistently improve over NF4, except for at very large block sizes.
This validates the claim that NF4 is not optimal at all block sizes.
However, at small block sizes their performance is very comparable, so this does not lead to a path to improving over the results of~\citet{qlora}.

The experiments are the cross product of the following:
\begin{enumerate}
    \item Models: LLaMA-7B/13B/30B~\citep{llama}, GPT-2-XL~\citep{gpt2}, GPT-Neo-2.7B~\citep{pile}.
    The latter two models are the HuggingFace Flax implementations~\citep{huggingface,flax} while the LLaMa models are reimplementations of the original using JAX + Haiku~\citep{jax,haiku}.
    \item Datasets: The WikiText-103 validation set~\citep{wikitext}, a subset of the PG-19 validation set~\citep{pg19}, the LAMBADA cloze prediction validation set~\citep{lambada}.
    \item Quantization methods: No quantization, NF4 quantization, AF4 quantization
\end{enumerate}
\textbf{Language modeling details.} In the interest of reducing the computation required, all experiments were run using disjoint inputs of length 512, rather than using the maximum attention context and a sliding window.
This leads to higher perplexities, but it is a fair setting for comparing degradation due to quantization.
Perplexities have been renormalized to be word-based rather than token-based using NLTK.

The LLaMA models were run using float16, and GPT-2 and GPT-Neo were run using bfloat16.
All experiments were run on one NVIDIA Quadro RTX 6000, except for the LLaMA-13B baselines and LLaMA-30B runs, which used an NVIDIA RTX A6000.

\textbf{Quantization details.} All of the models were quantized using a JAX transformation\footnote{This will be uploaded at \url{https://github.com/davisyoshida/abnormal-floats} as well}, so the model code was unmodified.
For matrices $W$ which are right-multiplied with the activations (i.e. $xW$), quantization is performed on column-wise blocks, while row-wise blocks are used for those which are left-multiplied\footnote{Of the models used here, only GPT-2 uses left multiplication}.

\subsection{Results}
\begin{figure*}[t!]
    \centering
    \includegraphics[width=0.95\textwidth]{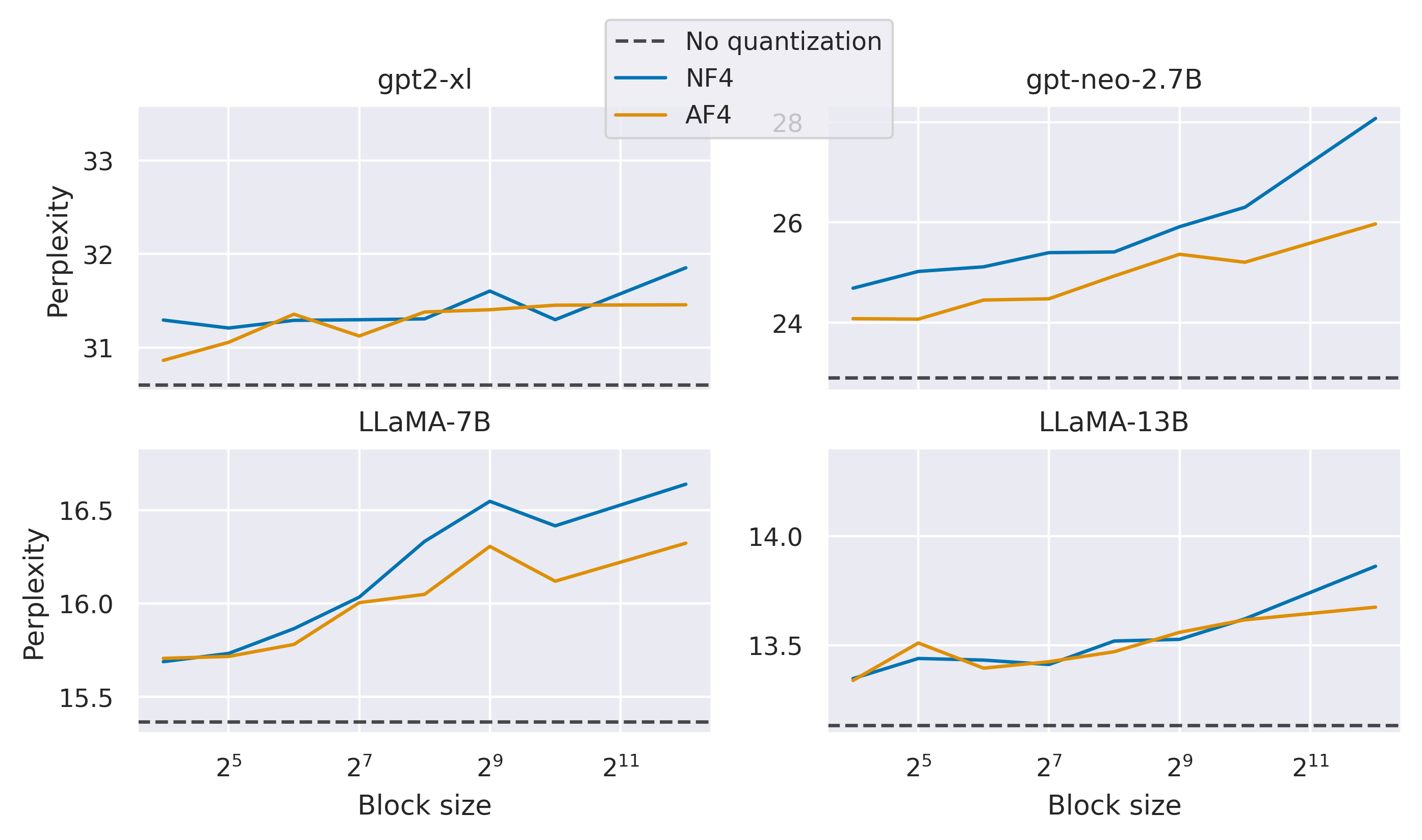}
    \caption{Perplexity on WikiText-103 validation set}\label{fig:wiki}
\end{figure*}

\begin{figure*}[hp!]
\centering
    \includegraphics[width=\textwidth]{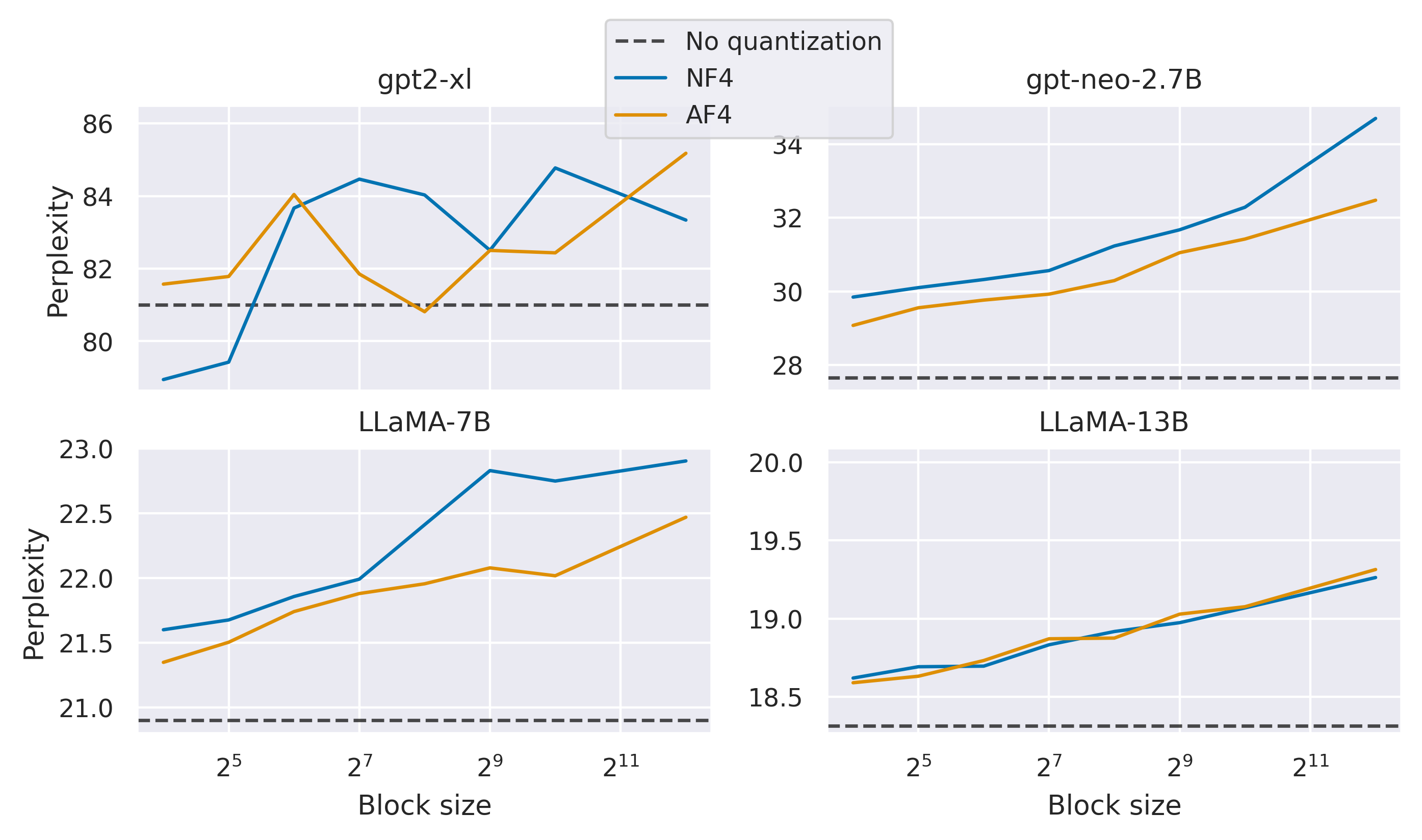}
    \caption{Perplexity on PG-19 validation set (subset)}\label{fig:pg19}
\end{figure*}

\begin{figure*}[hp!]
\centering
\begin{subfigure}{0.5\textwidth}
  \centering
  \includegraphics[width=\linewidth]{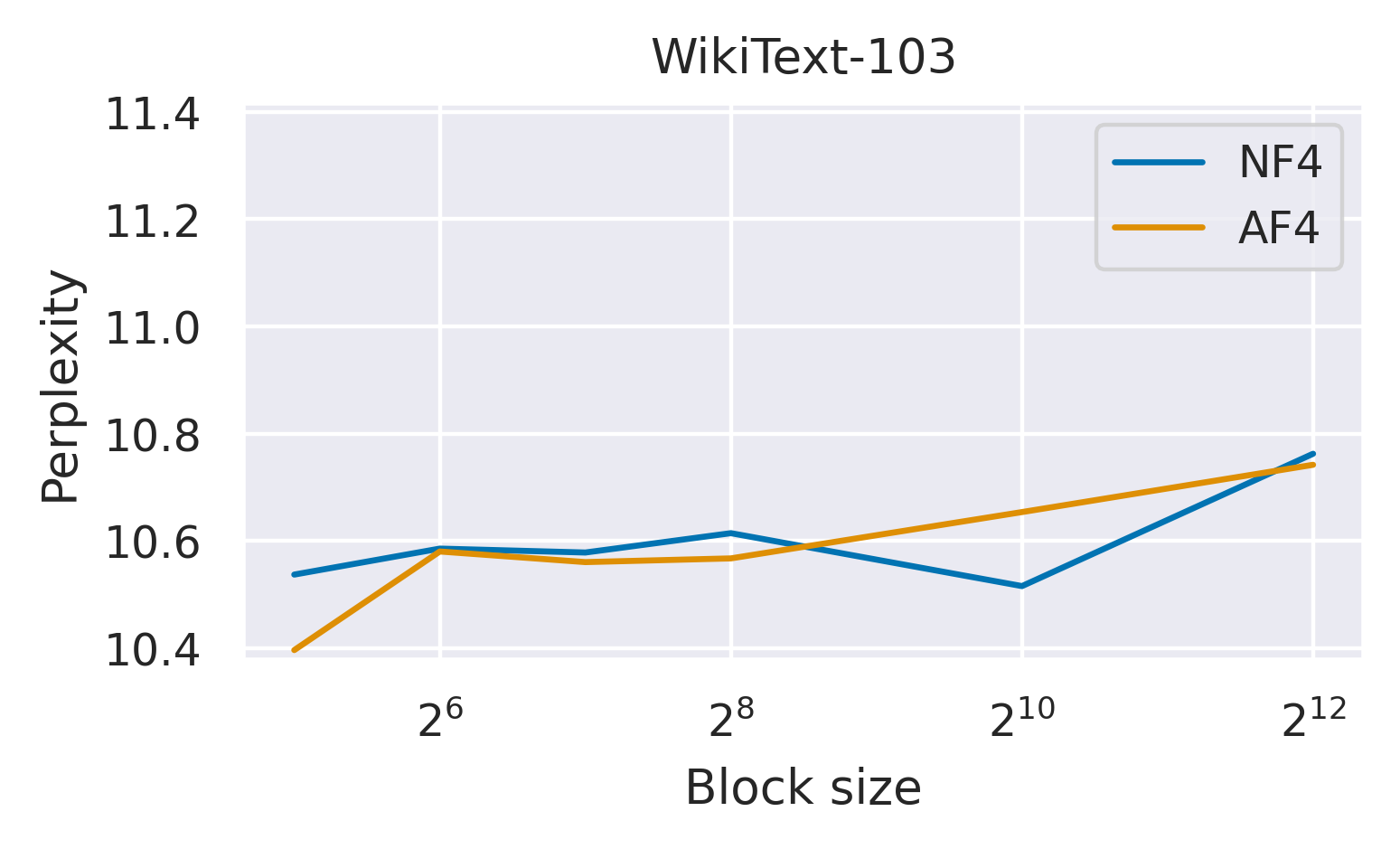}
\end{subfigure}%
\begin{subfigure}{.5\textwidth}
  \centering
  \includegraphics[width=\linewidth]{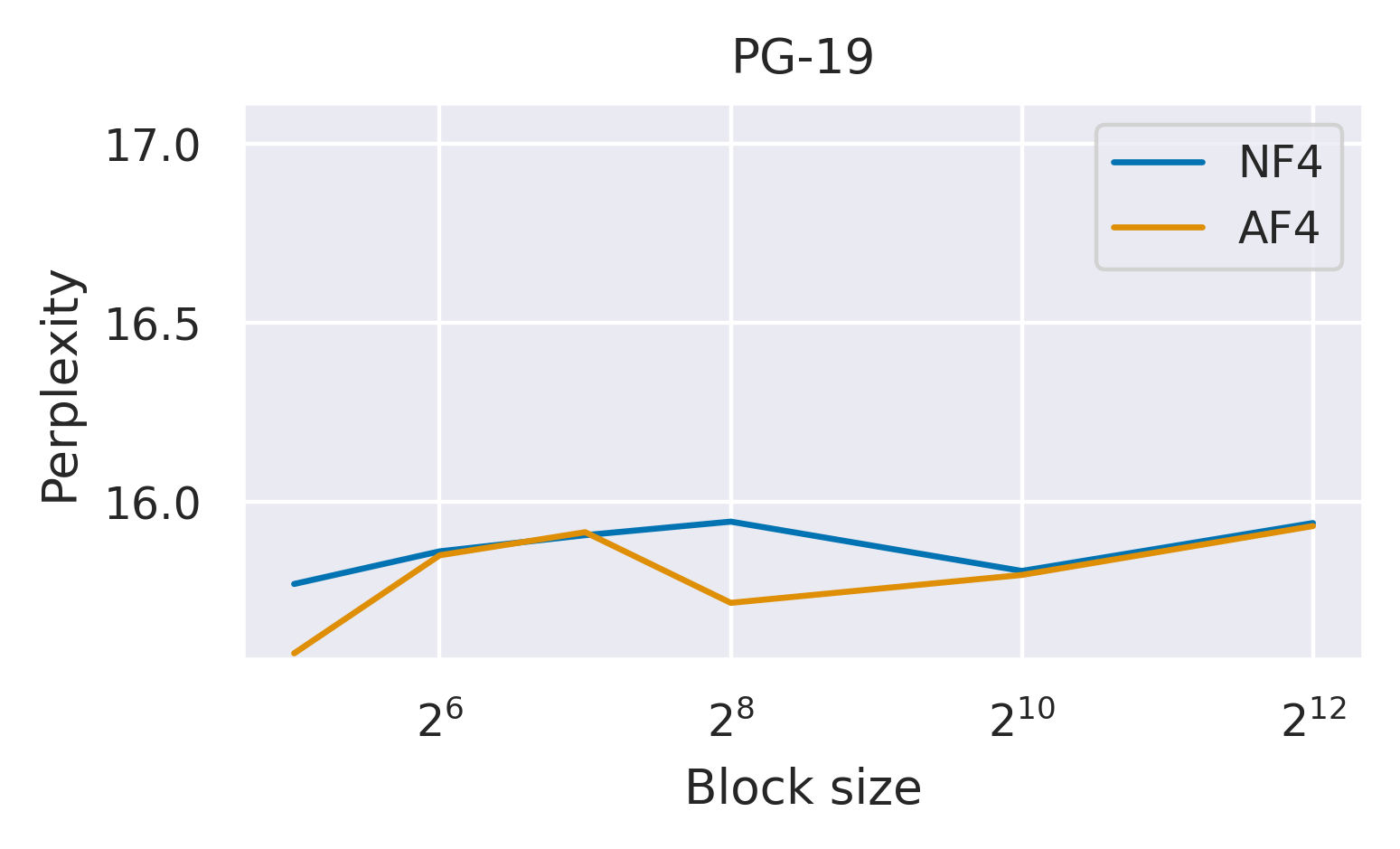}
\end{subfigure}
\caption{LLaMA-30B language modeling results}\label{fig:lm_30B}
\end{figure*}

\textbf{Language modeling}.
Results are shown in Figures~\ref{fig:wiki}, \ref{fig:pg19}, and \ref{fig:lm_30B}.
For 8 out of 10 model/dataset pairs except for GPT-2 and LLaMA-13B on PG-19, AF4 leads to at least slightly lower perplexity for a block size of $B = 4096$.
The LLaMA-13B/30B models show extremely little dependence on the choice of code, which is surprising given how different NF4 and AF4-4096 are.

For NF4's default block size of $B=64$, AF4 is only better in 6 out of 10 dataset/model pairs, but the only model it performs significantly better for at low block sizes is gpt-neo.

\begin{figure*}[hp!]
    \centering
    \includegraphics[width=\textwidth]{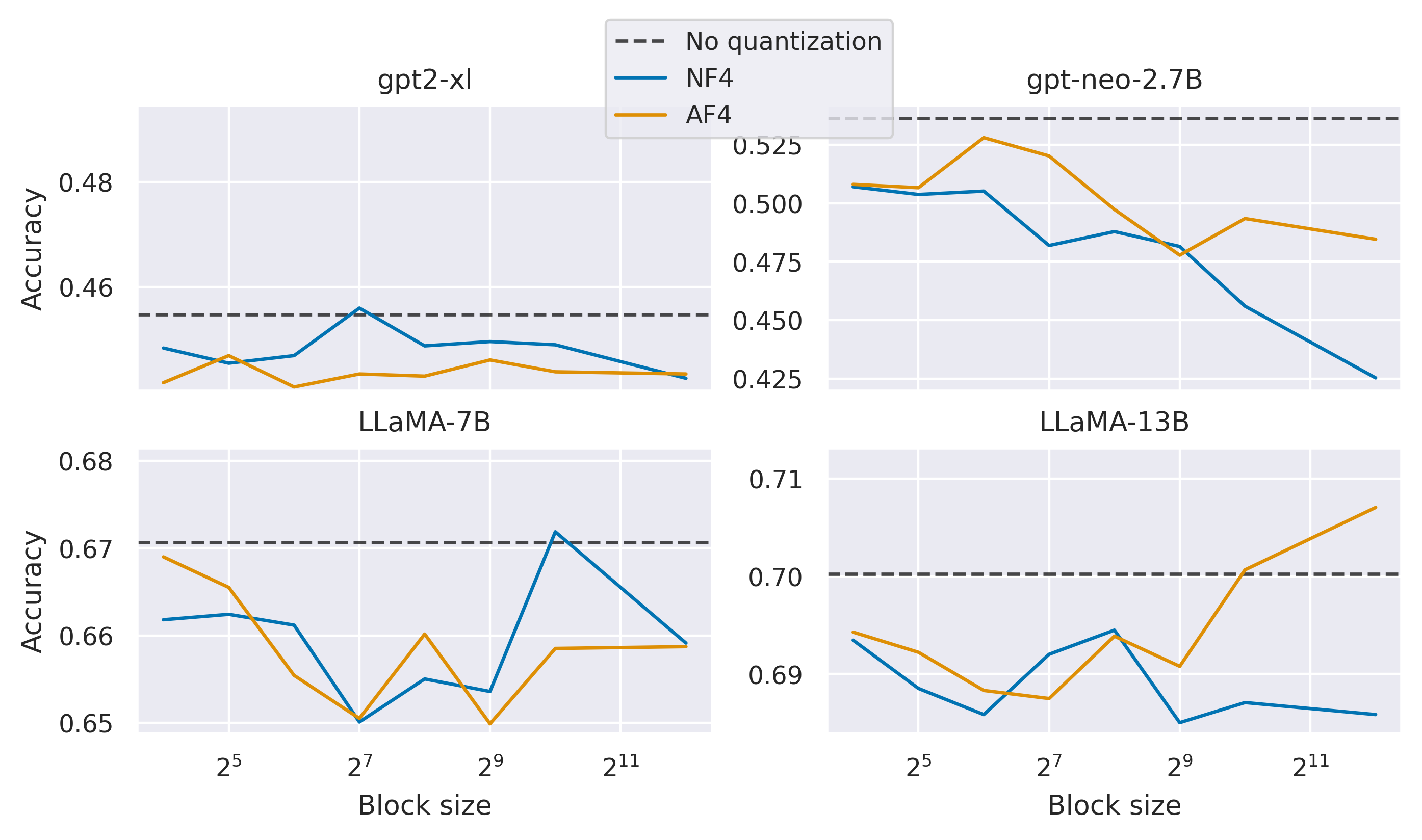}
    \caption{Accuracy on LAMBADA validation set}\label{fig:lambada}
\end{figure*}

\begin{figure}[ht!]
    \centering
    \includegraphics[width=\columnwidth]{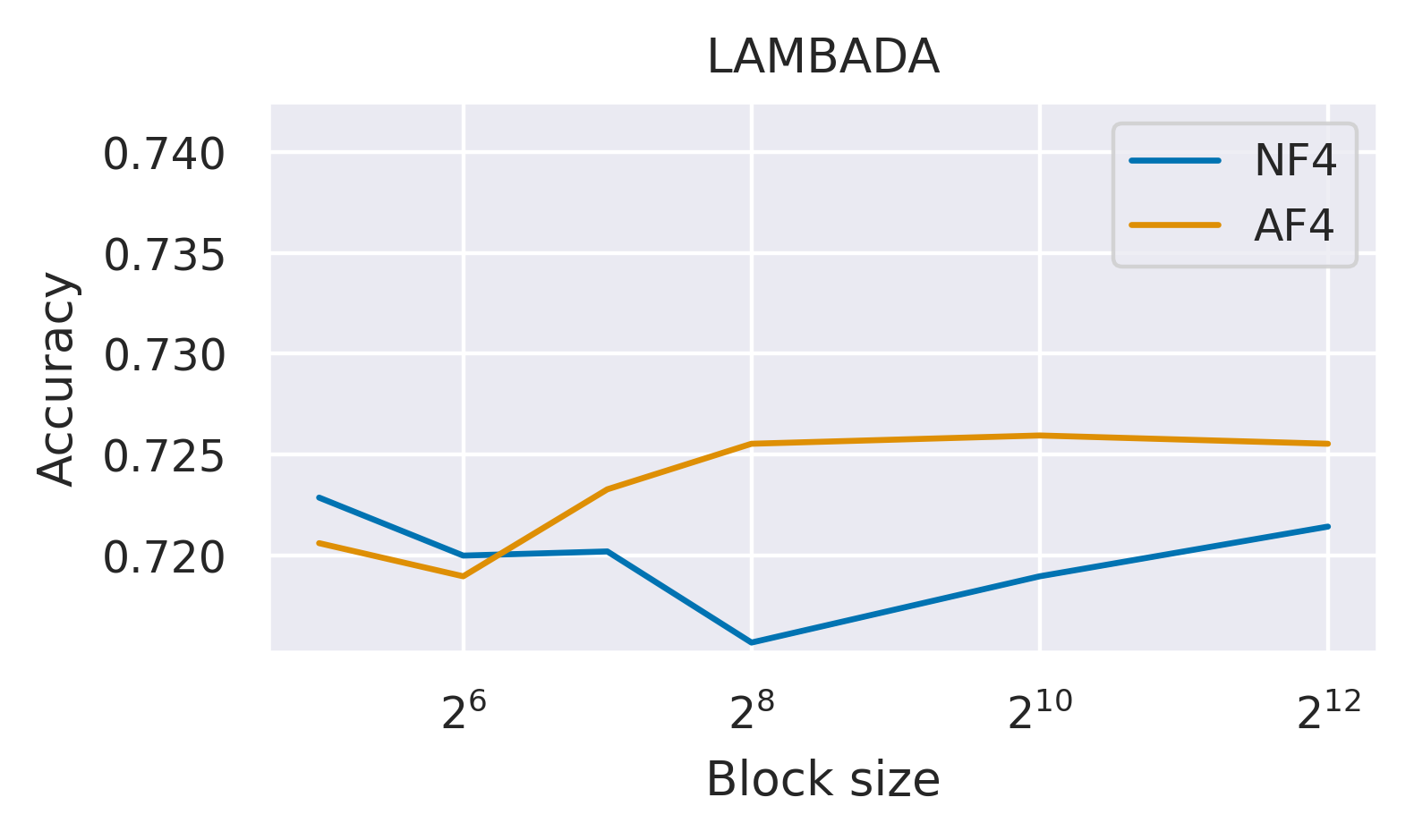}
    \caption{Accuracy of LLaMA-30B on LAMBADA validation set}\label{fig:lambada_30B}.
\end{figure}

\textbf{Zero-shot LAMBADA.} Accuracy on the LAMBADA validation set is shown in Figures~\ref{fig:lambada} and \ref{fig:lambada_30B}.
The results here seem extremely noisy.
For instance AF4-4096 leads to a better accuracy than FP16 for the LLaMA-13B model, which must be random chance rather than a meaningful measurement.

Similar to language modeling, there's some indication that AF4 may be better at large block sizes but even that is inconsistent.
NF4 leads to better accuracy with 3/5 models at its preferred block size of 64.

\subsection{Discussion}
A future investigation could use more models of size comparable to LLaMA-30B, and more classification datasets to attempt to average out the noisy results here.
However, these experiments do provide some evidence that it is possible to improve on NF4 for quantization at larger block sizes.
Since NF4 may be used at a small block size using the double quantization method of \citet{qlora}, this is mostly of theoretical interest.

As minimization of expected quantization error and uniformity of code value usage both fail to lead to codes which perform better across the board, it is an open question what criterion \emph{should} be used when selecting codes for data free quantization.

\section{Conclusion}
The main takeaway here should be that when doing block-wise quantization, the distribution of values to be quantized will depend on the block size.
This applies to quantization which uses a scale and an offset as well, although all the calculations here were for the scale-only case.

One also shouldn't try to aim for uniform usage of code values, since minimization of expected quantization error seems to be a better target.
The question of exactly why including the -1, 0, and 1 values is important despite worsening quantization error will hopefully be resolved by future work.

\section*{Acknowledgements}
Thank you to \texttt{sekstini} in the EleutherAI Discord for many productive conversations on this topic, and to Tim Dettmers for feedback on the first draft.

\newpage

\clearpage
\bibliography{bibliography}
\bibliographystyle{acl_natbib}

\FloatBarrier
\appendix
\section{Approximating the distribution of $X_i$}\label{approximation}
The following approximation of the continuous part of $X_i$'s distribution is based on a suggestion of user \texttt{whuber} on Cross Validated\footnote{\url{https://stats.stackexchange.com/questions/616752/does-the-following-distribution-converge-to-anything}}:
\begin{align}
    P&\left[ X_i \le  x \given[\middle] -1 < X_i < 1\right]\nonumber\\
    &\approx P\left[ Z_i < xm_0 \given[\middle] |Z_i| < m_0 \right]\nonumber\\
    &= \Psi(x m_0;m_0, 1)\label{eq:approx}
\end{align}
Where $m_0 = \halfnorm^{-1}\left(2^{-1/B}\right)$ and $\Psi(\cdot;a, \sigma)$ is a centered truncated normal distribution as before.
This is the same as the true CDF (Equation~\ref{eq:cdf}), but using a constant estimate of $M$ since it will concentrate around its median for large $B$.

\begin{figure}
    \centering
    \includegraphics[width=0.95\columnwidth]{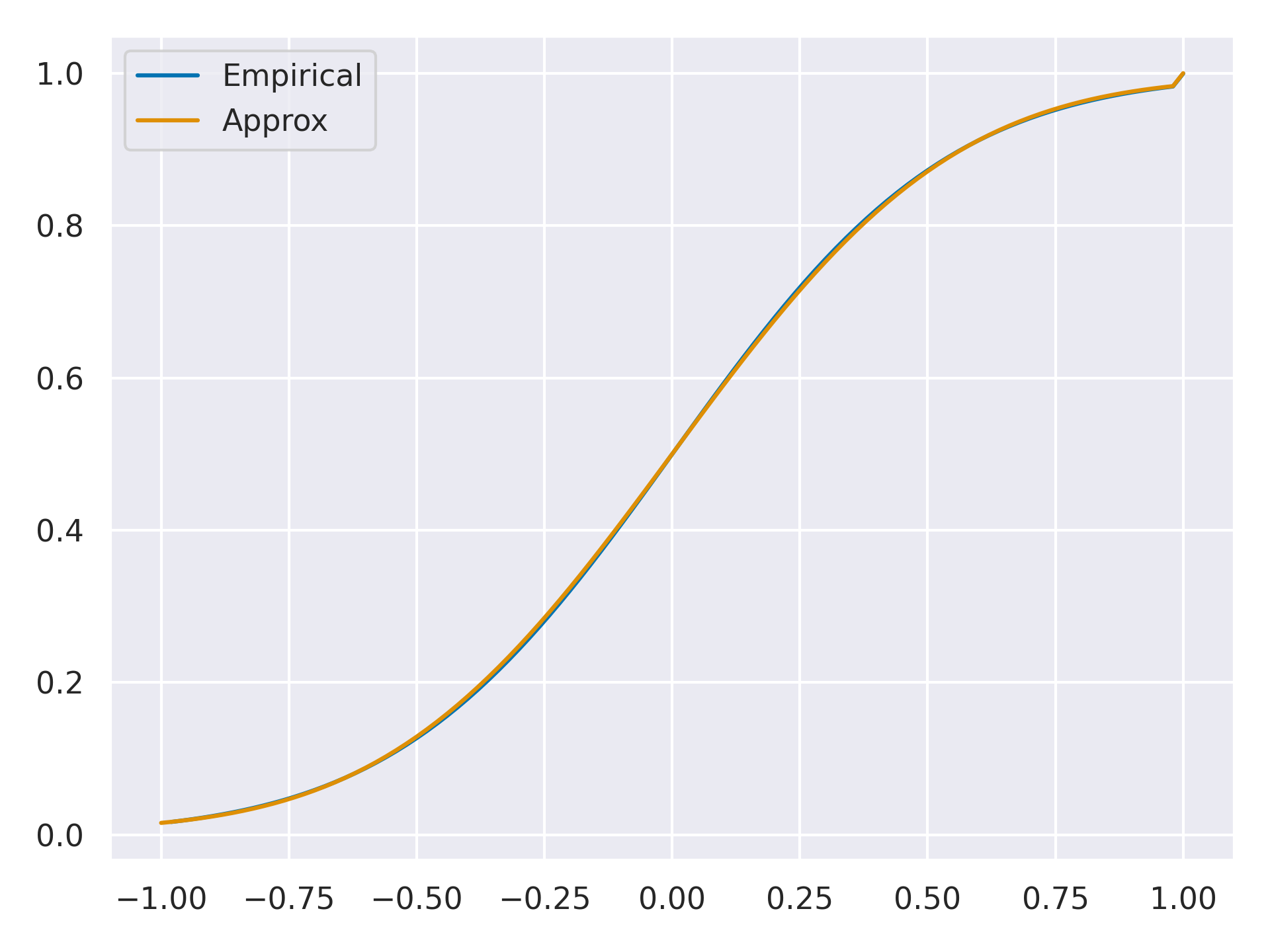}
    \caption{Comparison of empirical CDF of $X_i$ ($B=32$) and the approximation given in Equation~\ref{eq:approx}. $B=32$.}
    \label{fig:approx_cdf}
\end{figure}

Even for the smallest block size used here, $B=32$, this approximation appears to be quite close.
Figure~\ref{fig:approx_cdf} shows a comparison of the predicted CDF compared to an empirical CDF generated from $2^{33}$ samples\footnote{$2^{28}$ independent blocks of size 32}.
The empirical result and approximation are so close that I thought perhaps the integral worked out to just yield a truncated normal, but they turn out to differ slightly.
The approximate CDF predicts:
\begin{equation*}
P[X_i \le 1/2] \approx 0.8712
\end{equation*}
While drawing $2^{30}$ blocks of 32 and using one sample from each to estimate a 95\% confidence interval leads to:
\begin{equation*}
   P[X_i \le 1/2] \approx 0.8728 \pm 2 \times 10^{-5}
\end{equation*}

\section{Codes with uniform usage}\label{uniform}
\begin{figure}[h!]
    \centering
    \includegraphics[width=0.98\columnwidth]{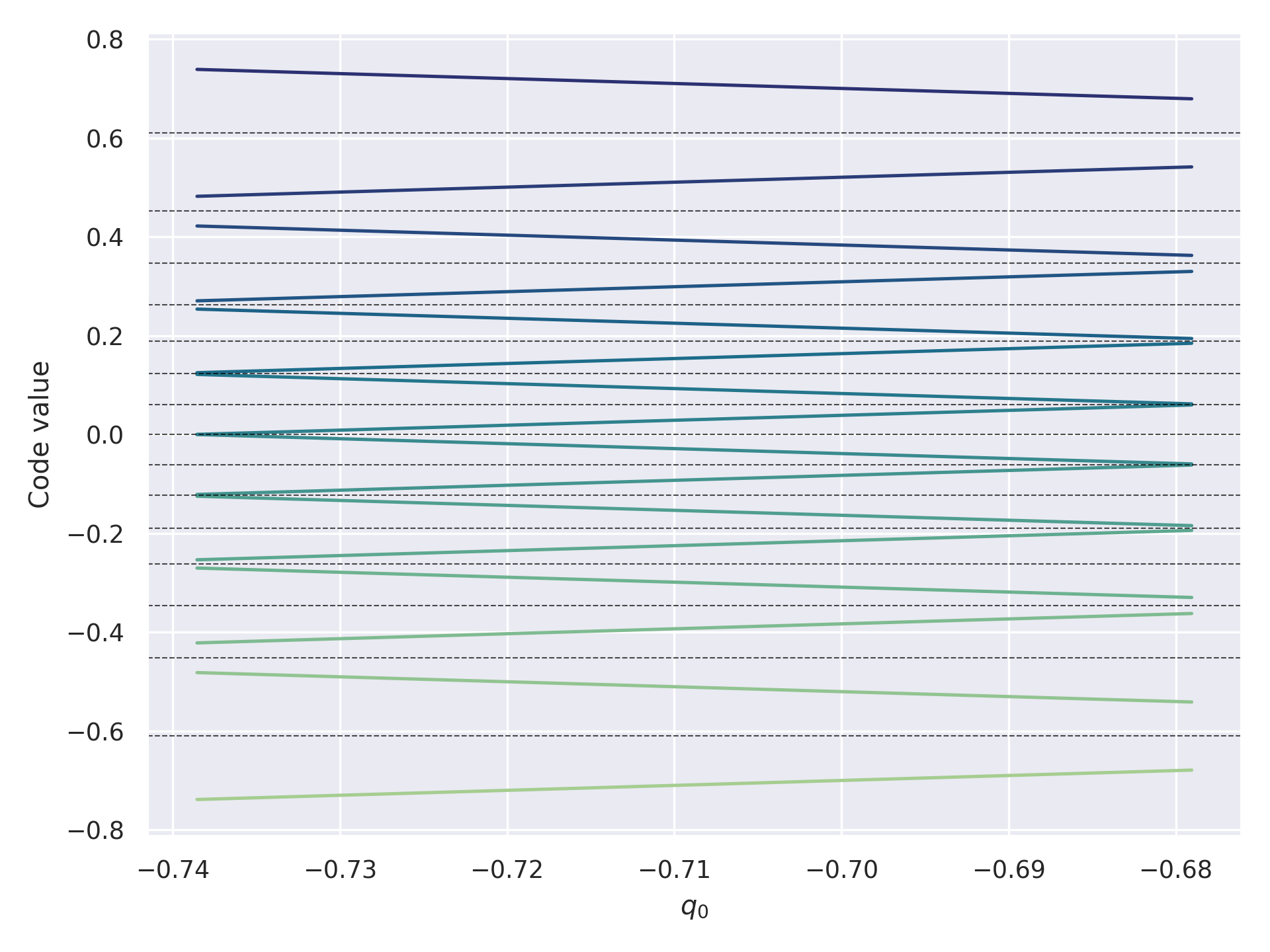}
    \caption{Each vertical slice represents a particular code which leads to uniform bin usage for $B=64$. The dashed lines are quantiles. Notice that in general the spacing is non-monotone with respect to distance from zero.}\label{fig:uniform_codes}
\end{figure}
To justify the parenthetical in the title, I'll briefly discuss what happens if one does try to optimize for uniform code value usage.
Using the method described in Section~\ref{construct_uniform}, it's easy to construct a set of codes with this property for all values of $B$.
As mentioned, the construction yields a variety of different codes depending on the selected $q_1$ value, which is illustrated in Figure~\ref{fig:uniform_codes}.

\begin{figure}[h]
    \centering
    \includegraphics[width=\columnwidth]{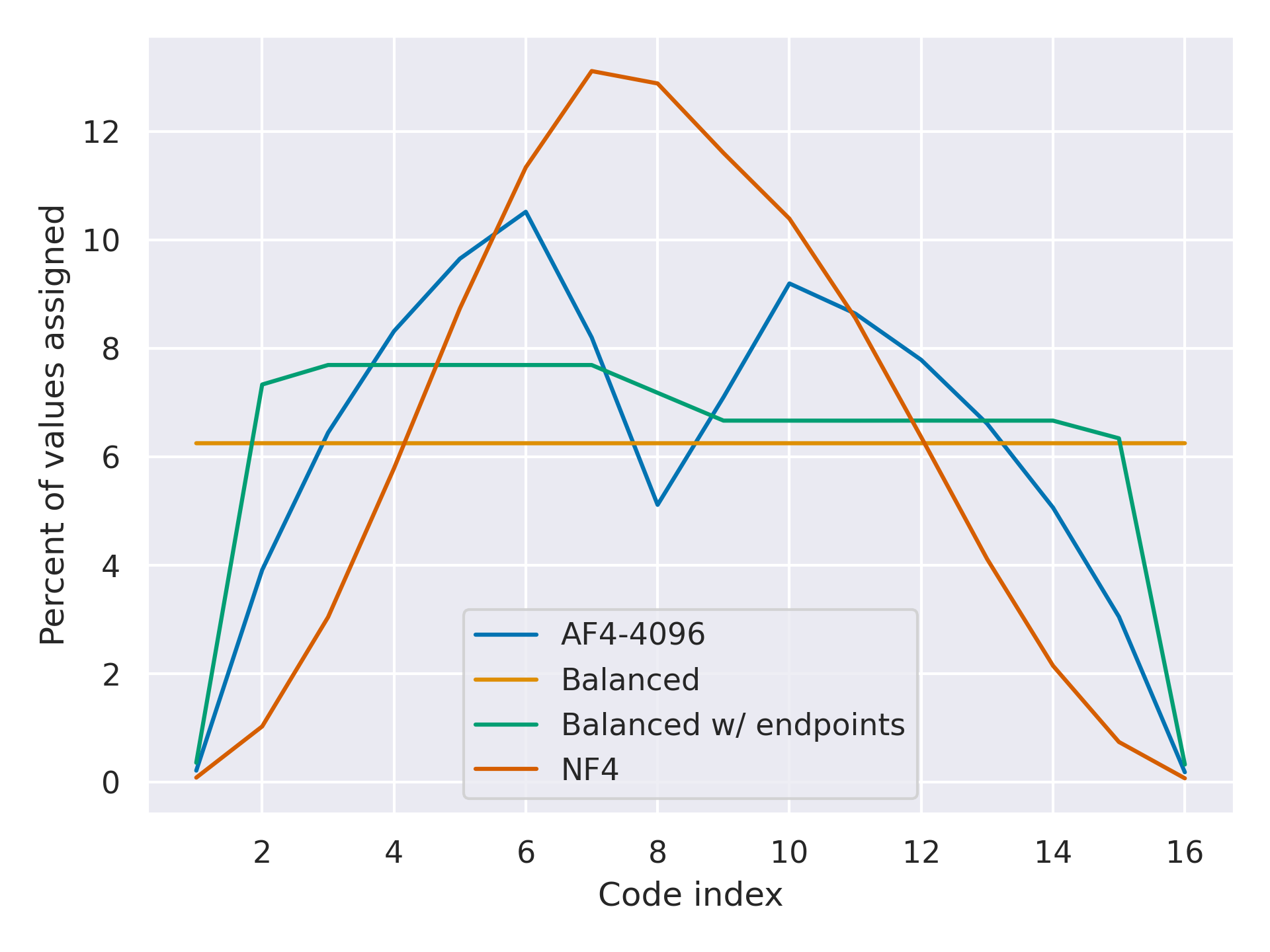}
    \caption{Usage of code values when absmax quantizing blocks of 4096 normal samples. ``Balanced'' shows successful construction of a code in which each value is used equally. ``Balanced w/ endpoints'' is a variant in which -1, 0, and 1 are included as code values.}\label{fig:uniform_code_usage}
\end{figure}
\begin{figure}[h!]
    \centering
    \includegraphics[width=\columnwidth]{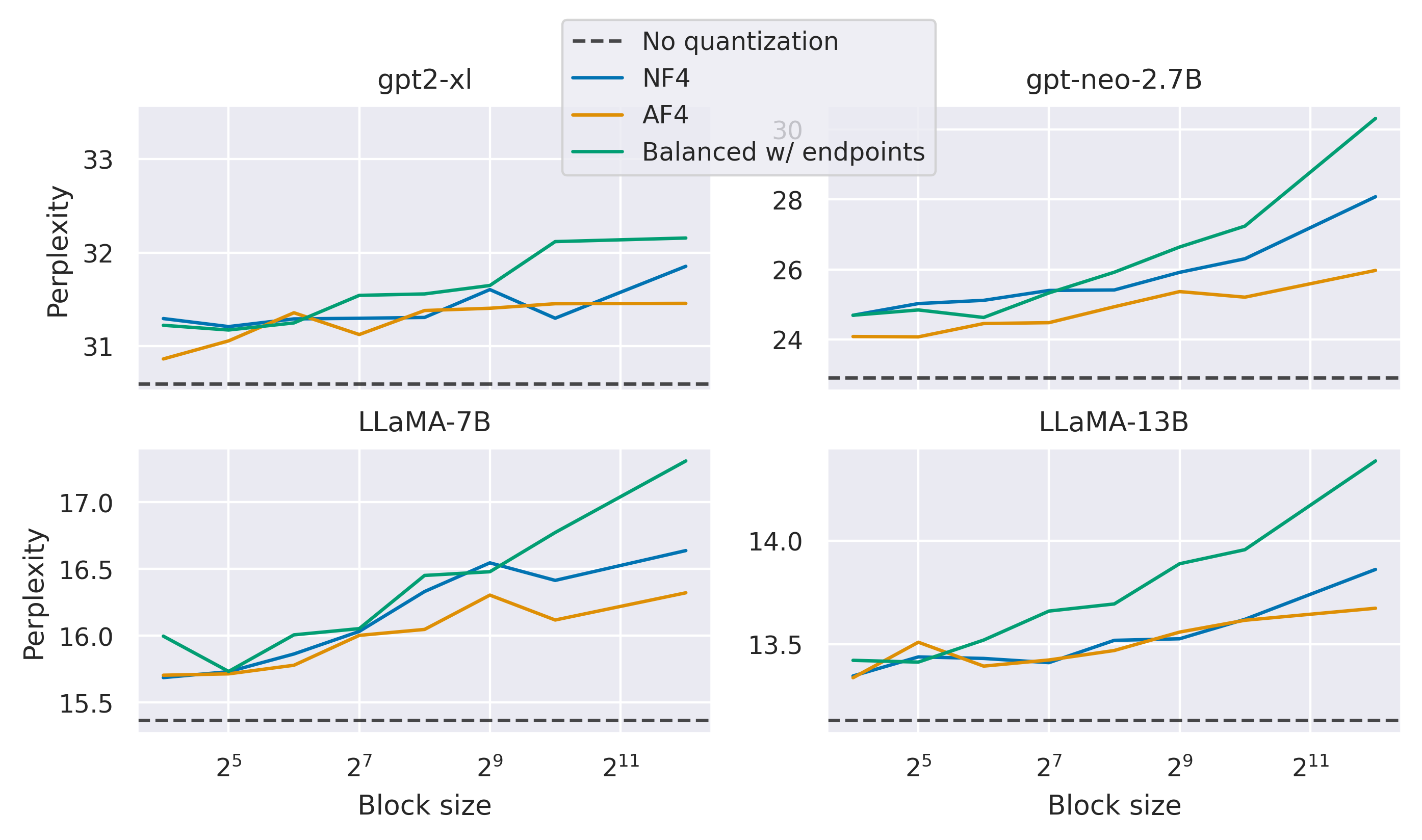}
    \caption{Perplexity on WikiText-103 validation set. Uniform code value usage does not lead to better quantization.}
    \label{fig:wikitext_balanced}
\end{figure}
Figure~\ref{fig:uniform_code_usage} compares the relative usage of code values between several different codes.
Despite the fact that ``Balanced w/ endpoints'' leads to less uniform usage of values than the exactly balanced code, including -1, 0, and 1 is necessary to achieve remotely acceptable quantization.
Figure~\ref{fig:wikitext_balanced} shows the perplexity resulting from quantization with this code.
It does about as well as NF4 and AF4 for small block sizes, but is much worse at large block sizes.
This clearly demonstrates that uniform code usage is not a sufficient code construction criterion.
\end{document}